\documentclass[10pt,twocolumn,letterpaper]{article}

\usepackage{iccv}
\usepackage{times}
\usepackage{epsfig}
\usepackage{graphicx}
\usepackage{amsmath}
\usepackage{amssymb}

\usepackage{subfigure}
\usepackage{algorithmic}
\usepackage{amsthm}
\usepackage{booktabs}

\usepackage[colorlinks,linkcolor=blue]{hyperref}
\usepackage[ruled,lined]{algorithm2e}

\usepackage{multirow}
\usepackage{multicol}
\usepackage{booktabs}
\newcommand{\tabincell}[2]{\begin{tabular}{@{}#1@{}}#2\end{tabular}}

\usepackage{color}

\iccvfinalcopy 

\def\iccvPaperID{****} 

\ificcvfinal\fi

\begin{document}


\title{Tokens-to-Token ViT: Training Vision Transformers from Scratch on ImageNet}

\author{Li Yuan\textsuperscript{\rm 1}\thanks{Work done during an internship at Yitu Tech.}, Yunpeng Chen\textsuperscript{\rm 2}, Tao Wang\textsuperscript{\rm 1,3}$^*$, Weihao Yu\textsuperscript{\rm 1}, Yujun Shi\textsuperscript{\rm 1}, \\Zihang Jiang\textsuperscript{\rm 1}, Francis E.H. Tay\textsuperscript{\rm 1}, Jiashi Feng\textsuperscript{\rm 1}, Shuicheng Yan\textsuperscript{\rm 1} \vspace{7pt}\\ 
\small{$^{1}$ National University of Singapore \quad}
\small{$^{2}$ YITU Technology \quad}
\small{$^{3}$ Institute of Data Science, National University of Singapore} \\

{\small\tt  yuanli@u.nus.edu},\,{\ \small\tt  yunpeng.chen@yitu-inc.com},\,{\ \small\tt  shuicheng.yan@gmail.com} 
}

\maketitle
\thispagestyle{empty}

\begin{abstract}
\vspace{-7pt}
Transformers, which are popular for language modeling, have been explored for solving vision tasks recently,  \eg, the Vision Transformer (ViT) for image classification. The ViT model splits each image into a sequence of tokens with fixed length and then applies multiple Transformer layers to model their global relation for classification. 
However, ViT achieves inferior performance to CNNs when trained from scratch on a midsize dataset like ImageNet.
We find it is because: 1) the simple tokenization of input images fails to model the important local structure such as edges and lines among neighboring pixels, leading to low training sample efficiency;
2) the redundant attention backbone design of ViT leads to limited feature richness for fixed computation budgets and limited training samples.
To overcome such limitations, we propose a new Tokens-To-Token Vision Transformer (T2T-ViT), which incorporates 1) a layer-wise Tokens-to-Token (T2T) transformation to progressively structurize the image to tokens by recursively aggregating neighboring Tokens into one Token (Tokens-to-Token), such that local structure represented by surrounding tokens can be modeled and tokens length can be reduced; 2) an efficient backbone with a deep-narrow structure for vision transformer motivated by CNN architecture design after empirical study. 
Notably, T2T-ViT reduces the parameter count and MACs of vanilla ViT by half, while achieving more than 3.0\% improvement when trained from scratch on ImageNet. It also outperforms ResNets and achieves comparable performance with MobileNets by directly training on ImageNet. 
For example, T2T-ViT with comparable size to ResNet50 (21.5M parameters) can achieve 83.3\% top1 accuracy in image resolution 384$\times$384 on ImageNet. \footnote{Code: https://github.com/yitu-opensource/T2T-ViT}
\end{abstract}
\vspace{-18pt}

\begin{figure}[t]
\begin{center}
\setlength{\tabcolsep}{1.5pt}
\includegraphics[scale=0.45]{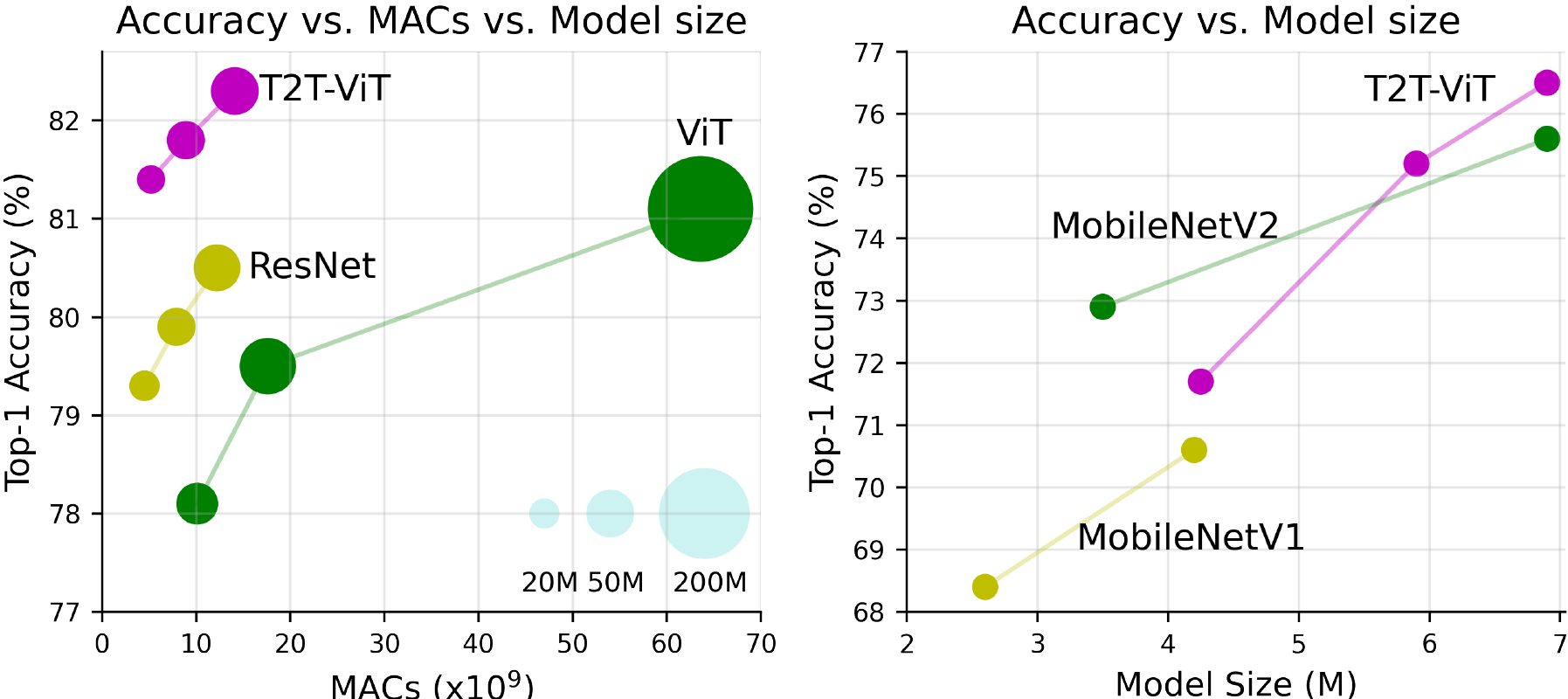}
\caption{Comparison between T2T-ViT with ViT, ResNets and MobileNets when trained from scratch on ImageNet. Left: performance curve of MACs vs. top-1 accuracy. Right: performance curve of model size vs. top-1 accuracy.}
\label{fig:perf_comparison}
\end{center}
\vspace{-26pt}
\end{figure}

\section{Introduction}

Self-attention models for language modeling like Transformers~\cite{vaswani2017attention} have been recently applied to vision tasks, including image classification~\cite{chen2020generative, dosovitskiy2020image, wu2020visual}, object detection~\cite{carion2020end,zhu2020deformable} and image processing like denoising, super-resolution and deraining~\cite{chen2020pre}.
Among them, the Vision Transformer (ViT)~\cite{dosovitskiy2020image} is the first full-transformer model that can be directly applied for image classification. 
In particular, ViT splits each image into $14\times14$ or $16\times16$ patches (\textit{a.k.a.}, tokens)  with fixed length; then following practice of the transformer for language modeling, ViT applies transformer layers to model the global relation among these tokens for classification.

\begin{figure*}[t]
\begin{center}
\setlength{\tabcolsep}{1.5pt}
\includegraphics[scale=0.5]{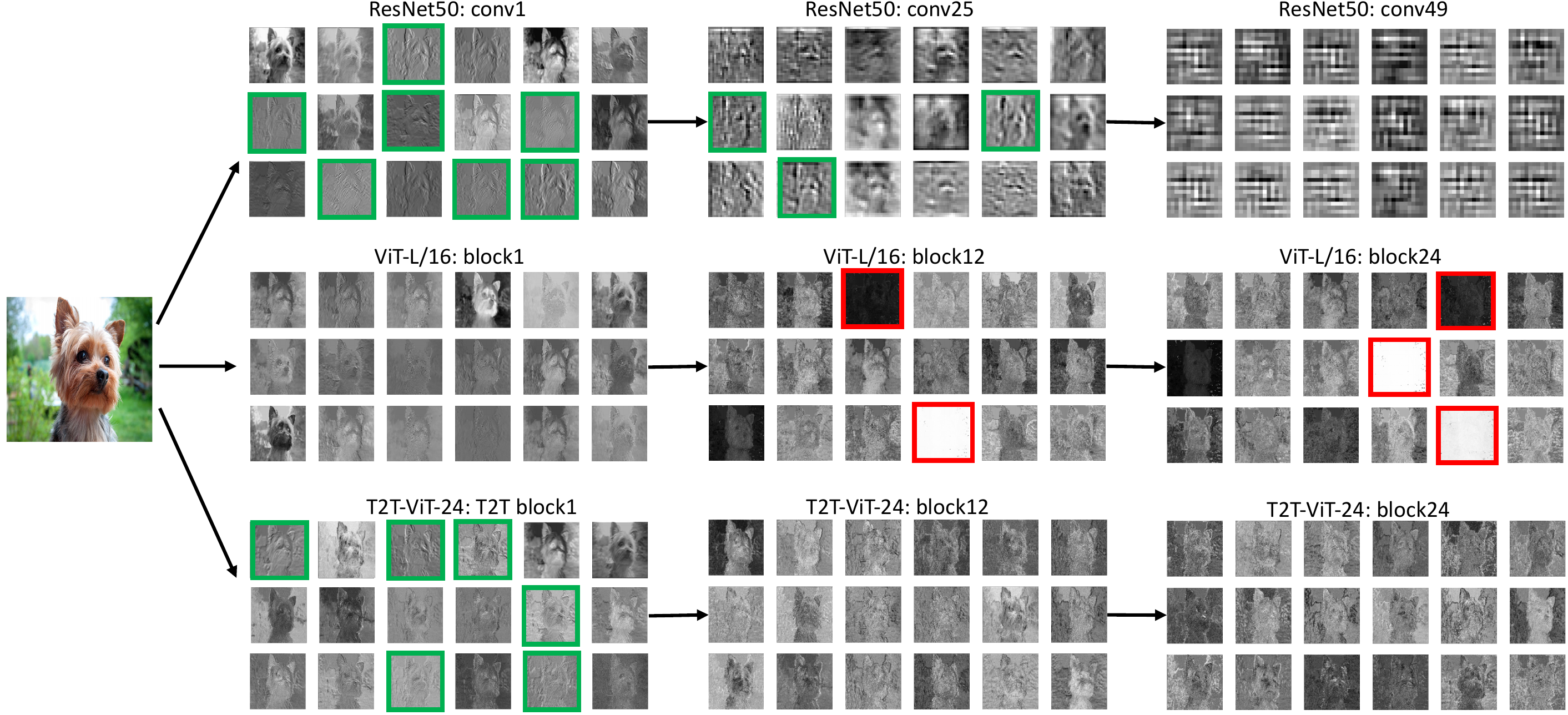}
\caption{Feature visualization of ResNet50, ViT-L/16~\cite{dosovitskiy2020image} and our proposed T2T-ViT-24 trained on ImageNet. Green boxes highlight learned low-level structure features such as edges and lines; red boxes highlight invalid feature maps with zero or too large values. Note the feature maps visualized here for ViT and T2T-ViT are not attention maps, but image features reshaped from tokens. For better visualization, we scale the input image to size $1024\times1024$ or $2048\times2048$.}
\label{fig:feature_visualization}
\end{center}
\vspace{-25pt}
\end{figure*}

Though ViT proves the full-transformer architecture is promising for vision tasks, its performance is still inferior to that of similar-sized CNN counterparts (\eg ResNets) when trained from scratch on a midsize dataset (\eg, ImageNet). 
We hypothesize that such performance gap roots in two main limitations of ViT: 
1) the straightforward tokenization of input images by hard split makes ViT unable to model the image local structure like edges and lines, and thus it requires significantly more training samples (like JFT-300M for pretraining) than CNNs for achieving similar performance; 
2) the attention backbone of ViT is not well-designed as CNNs for vision tasks, which contains redundancy and leads to limited feature richness and difficulties in model training.

To verify our hypotheses, we conduct a pilot study to investigate the difference in the learned features of ViT-L/16~\cite{dosovitskiy2020image} and ResNet50~\cite{he2016deep} through visualization in Fig.~\ref{fig:feature_visualization}. 
We observe the features of ResNet capture the desired local structure (edges, lines, textures, \etc) progressively from the bottom layer (conv1) to the middle layer (conv25).
However, the features of ViT are quite different: the structure information is poorly modeled while the global relations (\eg, the whole dog) are captured by all the attention blocks.
These observations indicate that the vanilla ViT ignores the local structure when directly splitting images to tokens with fixed length. Besides, we find many channels in ViT have zero value (highlighted in red in Fig.~\ref{fig:feature_visualization}), implying the backbone of ViT is not efficient as ResNets and offers limited feature richness when training samples are not enough.
\vspace{-5pt}

We are then motivated to design a new full-transformer vision model to overcome above limitations. 
1) Instead of the naive tokenization used in ViT~\cite{dosovitskiy2020image}, we propose a progressive tokenization module to aggregate neighboring \textit{Tokens} to one \textit{Token} (named Tokens-to-Token module), which can model the local structure information of surrounding tokens and reduce the length of tokens iteratively.
Specifically, in each Token-to-Token (T2T) step, the tokens output by a transformer layer are reconstructed as an image (\textit{re-structurization}) which is then split into tokens with overlapping (\textit{soft split}) and finally the surrounding tokens are aggregated together by flattening the split patches.
Thus the local structure from surrounding patches is embedded into the tokens to be input into the next transformer layer.
By conducting T2T iteratively, the local structure is aggregated into tokens and the length of tokens can be reduced by the aggregation process. 
2) To find an efficient backbone for vision transformers, we explore borrowing some architecture designs from CNNs to build transformer layers for improving the feature richness, and we find  ``deep-narrow'' architecture design with fewer channels but more layers in ViT brings much better performance at comparable model size and MACs (Multi-Adds).
Specifically, we investigate Wide-ResNets (shallow-wide vs deep-narrow structure)~\cite{zagoruyko2016wide}, DenseNet  (dense connection)~\cite{huang2017densely}, ResneXt structure~\cite{xie2017aggregated}, Ghost operation~\cite{han2020ghostnet, zhou2019neural} and channel attention~\cite{hu2018squeeze}.
We find among them, deep-narrow structure~\cite{zagoruyko2016wide} is the most efficient and effective for ViT, reducing the parameter count and MACs significantly with nearly no degradation in performance.
This also indicates the architecture engineering of CNNs can benefit the backbone design of vision transformers.

Based on the T2T module and deep-narrow backbone architecture, we develop the Tokens-to-Token Vision Transformer (T2T-ViT), which significantly boosts the performance when trained from scratch on ImageNet (Fig.~\ref{fig:perf_comparison}), and is more lightweight than the vanilla ViT.
As shown in Fig.~\ref{fig:perf_comparison}, our T2T-ViT with 21.5M parameters and 4.8G MACs can achieve 81.5\% top-1 accuracy on ImageNet, much higher than that of ViT~\cite{dosovitskiy2020image} with 48.6M parameters and 10.1G MACs (78.1\%). This result is also higher than the popular CNNs of similar size, like ResNet50 with 25.5M parameters (76\%-79\%). 
Besides, we also design lite variants of T2T-ViT by simply adopting fewer layers, which achieve comparable results with MobileNets~\cite{howard2017mobilenets, sandler2018mobilenetv2} (Fig.~\ref{fig:perf_comparison}).

To sum up, our contributions are three-fold:

\begin{itemize}
\item For the first time, we show by carefully designing transformers architecture (T2T module and efficient backbone), visual transformers can outperform CNNs at different complexities on ImageNet without pre-training on JFT-300M.
\vspace{-5pt}
\item We develop a novel progressive tokenization for ViT and demonstrate its advantage over the simple tokenization approach by ViT, and we propose a T2T module that can encode the important local structure for each token.
\vspace{-5pt}
\item We show the architecture engineering of CNNs can benefit the backbone design of ViT to improve the feature richness and reduce redundancy. Through extensive experiments, we find deep-narrow architecture design works best for ViT.
\vspace{-5pt}

\end{itemize}

\section{Related Work}
\vspace{-5pt}
\paragraph{Transformers in Vision}
Transformers~\cite{vaswani2017attention} are the models that entirely rely on the self-attention mechanism to draw global dependencies between input and output, and currently they have dominated natural language modelling~\cite{devlin2018bert,radford2018improving,brown2020language,yang2019xlnet,peters2019knowledge,liu2019roberta}.
A transformer layer usually consists of a multi-head self-attention layer (MSA) and an MLP block. 
Layernorm (LN) is applied before each layer and residual connections in both the self-attention layer and MLP block.
Recent works have explored applying transformers to various vision tasks: image classification~\cite{chen2020generative, dosovitskiy2020image}, object detection~\cite{carion2020end,zhu2020deformable,zheng2020end,dai2020up,sun2020rethinking}, segmentation~\cite{chen2020pre,wang2020end}, image enhancement~\cite{chen2020pre,yang2020learning}, image generation~\cite{parmar2018image}, video processing~\cite{zhou2018end,zeng2020learning}, and 3D point cloud processing~\cite{zhao2020point}. 
Among them, the Vision Transformer (ViT) proves that a pure Transformer architecture can also attain state-of-the-art performance on image classification. 
However, ViT heavily relies on large-scale datasets such as ImageNet-21k and JFT-300M (which is not publically available) for model pretraining, requiring huge computation resources. 
In contrast, our proposed T2T-ViT is more efficient and can be trained on ImageNet without using those large-scale datasets. 
A recent concurrent work DeiT~\cite{touvron2020training} applies Knowledge Distillation~\cite{hinton2015distilling, yuan2020revisiting} to improve the original ViT by adding a KD token along with the class token, which is orthogonal to our work,  as our T2T-ViT focuses on the architecture design, and our T2T-ViT can achieve higher performance than DeiT without CNN as teacher model.  

\vspace{-12pt}
\paragraph{Self-attention in CNNs} Self-attention mechanism has been widely applied to CNNs in vision task~\cite{wang2017residual, zhao2018psanet, hu2018gather,yang2016stacked, hu2018squeeze, wang2018non, bello2019attention, chen20182, hu2019local,ramachandran2019stand, woo2018cbam, fu2019dual, yuan2020toward, yuan2020simple}. 
Among these works, the SE block~\cite{hu2018squeeze} applies attention to channel dimensions and non-local networks~\cite{wang2018non} are designed for capturing long-range dependencies via global attention.
Compared with most of the works exploring global attention on images~\cite{bello2019attention, woo2018cbam, fu2019dual, wang2018non}, some works~\cite{hu2019local, ramachandran2019stand} also explore self-attention in a local patch to reduce the memory and computation cost.
More recently, SAN~\cite{zhao2020exploring} investigates both pairwise and patchwise self-attention for image recognition, where the patchwise self-attention is a generalization of convolution.
In this work, we also replace the T2T module with multiple convolution layers in experiments and find the convolution layers do not perform better than our designed T2T module. 
\vspace{-3pt}

\section{Tokens-to-Token ViT}
To overcome the limitations of simple tokenization and inefficient backbone of ViT, we propose Tokens-to-Token Vision Transformer (T2T-ViT) which can progressively tokenize the image to tokens and has an efficient backbone.
Hence, T2T-ViT consists of two main components (Fig.~\ref{fig:network_structure}):
1) a layer-wise ``Tokens-to-Token module'' (T2T module) to model the local structure information of the image and reduce the length of tokens progressively; 
2) an efficient ``T2T-ViT backbone'' to draw the global attention relation on tokens from the T2T module.
We adopt a deep-narrow structure for the backbone to reduce redundancy and improve the feature richness after exploring several CNN-based architecture designs. 
We now explain these components one by one. 

\subsection{Tokens-to-Token: Progressive Tokenization} 
\label{sec:method_T2T}
\begin{figure}
\begin{center}
\setlength{\tabcolsep}{1.5pt}
\includegraphics[scale=0.45]{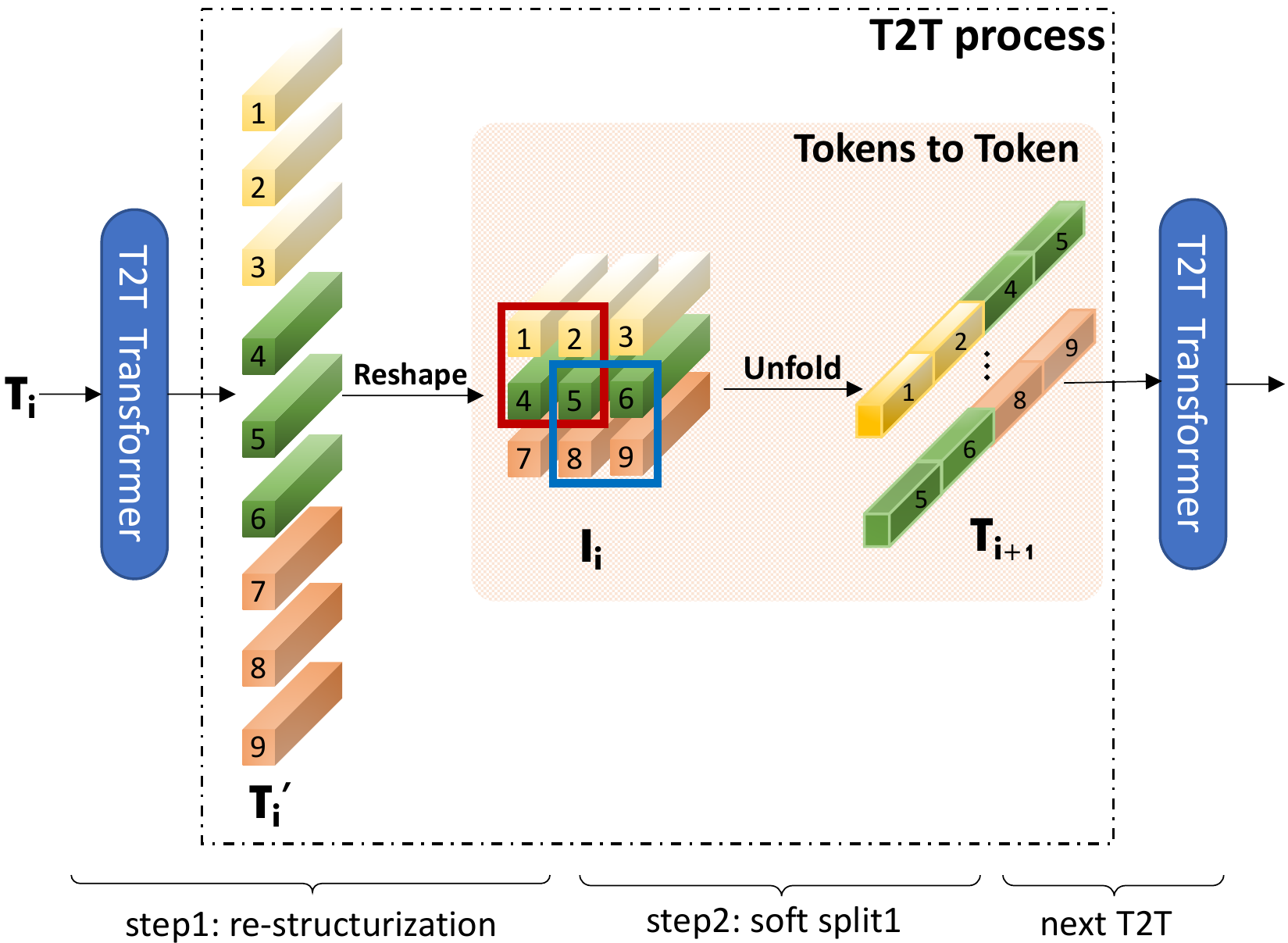}
\caption{Illustration of T2T process. 
The tokens $T_i$ are re-structurized as an image $I_i$ after transformation and reshaping; then $I_i$ is split with overlapping to tokens $T_{i+1}$ again.
Specifically, as shown in the pink panel, the four tokens (1,2,4,5) of the input $I_i$ are concatenated to form one token in $T_{i+1}$.
The T2T transformer can be a normal Transformer layer~\cite{vaswani2017attention} or other efficient transformers like Performer layer~\cite{sun2020rethinking} at limited GPU memory. }
\label{fig:t2t}
\end{center}
\vspace{-22pt}
\end{figure}

The Token-to-Token (T2T) module aims to overcome the limitation of simple tokenization in ViT.
It progressively structurizes an image to tokens and models the local structure information, and in this way the length of tokens can be reduced iteratively. 
Each T2T process has two steps: \textit{Re-structurization}  and  \textit{Soft Split (SS) } (Fig.~\ref{fig:t2t}). 
\vspace{-10pt}

\paragraph{Re-structurization}  As shown in Fig.~\ref{fig:t2t}, given a sequence of tokens $T$ from the preceding transformer layer, it will be transformed by the self-attention block (the T2T transformer in Fig.~\ref{fig:t2t}):
\begin{equation}
    T' = \text{MLP}(\text{MSA}(T)),
\end{equation}
where MSA denotes the multihead self-attention operation with layer normalization and ``MLP'' is the multilayer perceptron with layer normalization in the standard Transformer~\cite{dosovitskiy2020image}. 
Then the tokens $ T' $ will be reshaped as an image in the spatial dimension,
\begin{equation}
    I = \text{Reshape}(T').
\end{equation}
Here ``Reshape'' re-organizes tokens $T'\in \mathbb{R}^{l\times c}$ to $I\in \mathbb{R}^{h\times w\times c}$, where $l$ is the length of $T'$, $h$, $w$, $c$ are height, width and channel respectively, and $l = h\times w$.

\vspace{-10pt}
\paragraph{Soft Split} As shown in Fig.~\ref{fig:t2t}, after obtaining the re-structurized image $I$, we apply the soft split on it to model local structure information and reduce length of tokens. 
Specifically, to avoid information loss in generating tokens from the re-structurizated image, we split it into patches with overlapping. 
As such, each patch is correlated with surrounding patches to establish a prior that there should be stronger correlations between surrounding tokens. 
The tokens in each split patch are concatenated as one token (Tokens-to-Token, Fig.~\ref{fig:t2t}), and thus the local information can be aggregated from surrounding pixels and patches. 

When conducting the soft split, the size of each patch is $k\times k$ with $s$ overlapping and $p$ padding on the image,
where $k-s$ is similar to the stride in convolution operation.
So for the reconstructed image $I\in \mathbb{R}^{h\times w \times c}$, the length of output tokens $T_{o}$ after soft split is
\begin{equation}
    l_o = \left \lfloor \frac{h +2p-k}{k-s} + 1 \right \rfloor \times \left \lfloor \frac{w +2p-k}{k-s} + 1 \right \rfloor.
\label{eqn:solf_split}
\end{equation}
Each split patch has size $k\times k\times c$. We flatten all patches in spatial dimensions to tokens $T_{o}\in \mathbb{R}^{l_o\times ck^2}$.  After the soft split, the output tokens are fed for the next T2T process. 

\vspace{-10pt}
\paragraph{T2T module} By conducting the above Re-structurization and Soft Split iteratively, the T2T module can progressively reduce the length of tokens and transform the spatial structure of the image. The iterative process in T2T module can be formulated as
\begin{equation}
\begin{aligned}
 &T'_{i} = \text{MLP}(\text{MSA}(T_{i}),  \\
 &I_{i} = \text{Reshape}(T'_{i}) , \\
 &T_{i+1} = \text{SS}(I_i), & i=1...(n-1).
\end{aligned}
\end{equation}
For the input image $I_0$, we apply a soft split at first to split it to tokens: $T_1=\text{SS}(I_0)$. 
After the final iteration, the output tokens $T_f$ of the T2T module has fixed length, so the backbone of T2T-ViT can model the global relation on $T_f$.

Additionally, as the length of tokens in the T2T module is larger than the normal case ($16\times16$) in ViT, the MACs and memory usage are huge.
To address the limitations, in our T2T module, we set the channel dimension of the T2T layer small (32 or 64) to reduce MACs, and optionally adopt an efficient Transformer such as Performer~\cite{choromanski2020rethinking} layer to reduce memory usage at limited GPU memory.
We provide an ablation study on the difference between adopting standard Transformer layer and Performer layer in our experiments.


\subsection{T2T-ViT Backbone}
\label{sec:backbone}

\begin{figure*}[h!]
\begin{center}
\setlength{\tabcolsep}{1.5pt}
\includegraphics[scale=0.55]{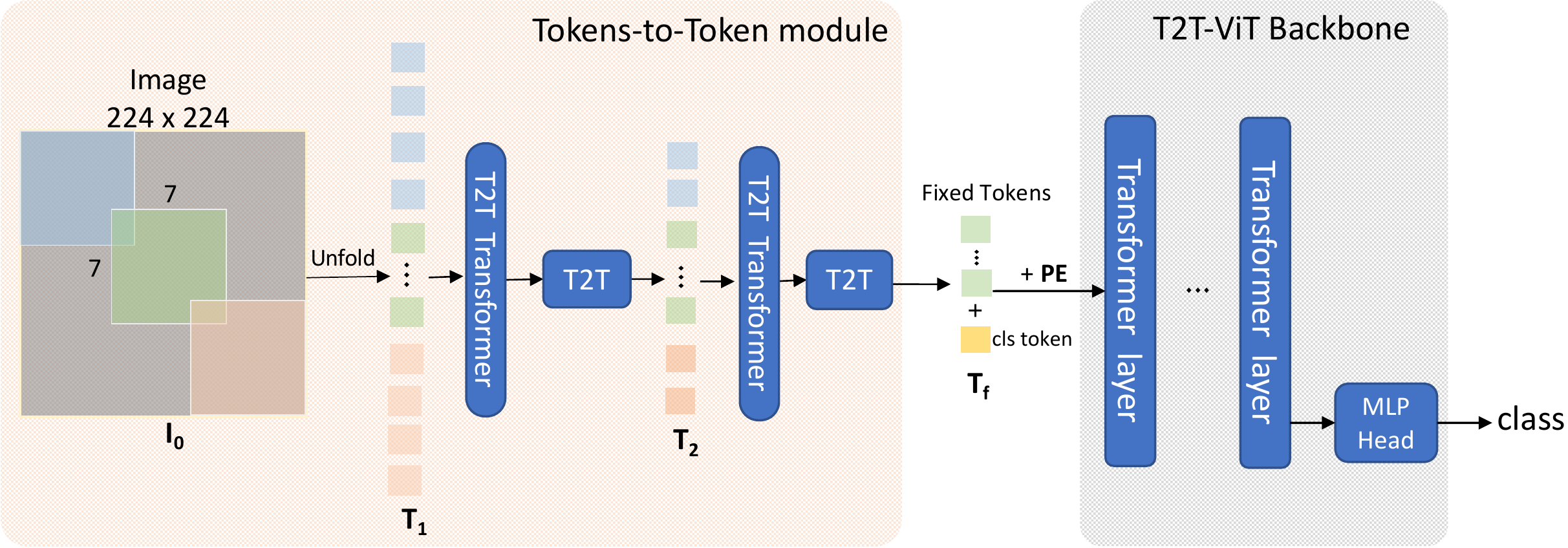}
\caption{The overall network architecture of T2T-ViT. 
In the T2T module, the input image is first soft split as patches, and then unfolded as a sequence of tokens $T_0$.  
The length of tokens is reduced progressively in the T2T module (we use two iterations here and output $T_f$). 
Then the T2T-ViT backbone takes the fixed tokens as input and outputs the predictions. The two T2T blocks are the same as Fig.~\ref{fig:t2t} and PE is Position Embedding. }
\label{fig:network_structure}
\end{center}
\vspace{-15pt}
\end{figure*}

\begin{table*}
\small
\begin{center}
\fontsize{8pt}{12pt}\selectfont
\setlength\tabcolsep{5pt}
\caption{Structure details of  T2T-ViT.
T2T-ViT-14/19/24 have comparable model size with ResNet50/101/152. T2T-ViT-7/12 have comparable model size with MobileNetV1/V2.
For T2T transformer layer, we adopt Transformer layer for T2T-ViT$_t$-14 and Performer layer for T2T-ViT-14 at limited GPU memory. 
For ViT, `S' means Small, `B' is Base and `L' is Large.
`ViT-S/16' is a variant from original ViT-B/16~\cite{dosovitskiy2020image} with smaller MLP size and layer depth.}
\begin{tabular}{l|cccc|ccc|cc}
\toprule
\multirow{3}{*}{Models} & \multicolumn{4}{c|}{Tokens-to-Token module} & \multicolumn{3}{c|}{T2T-ViT backbone} & \multicolumn{2}{c}{Model size} \\
\cline{2-10}
~ &\tabincell{c}{T2T\\transformer} &Depth & \tabincell{c}{Hidden\\dim} & \tabincell{c}{MLP\\size}& Depth &\tabincell{c}{Hidden\\dim} &\tabincell{c}{MLP\\size}  & \tabincell{c}{Params\\(M)} &\tabincell{c}{MACs\\(G)}\\
 \midrule
 ViT-S/16~\cite{dosovitskiy2020image}  &-&-&-&-&8  &786  &2358 &48.6 & 10.1 \\
 ViT-B/16~\cite{dosovitskiy2020image} &-&-&-&-&12 &786  &3072 &86.8 & 17.6 \\
 ViT-L/16~\cite{dosovitskiy2020image} &-&-&-&-&24 &1024 &4096 &304.3 & 63.6\\
 \midrule
  T2T-ViT-14 &Performer&2&64&64&14 &384 &1152  & 21.5 & 4.8\\
  T2T-ViT-19 &Performer&2&64&64&19 &448 &1344 &39.2 &8.5  \\
  T2T-ViT-24 &Performer&2&64&64&24 &512 &1536 &64.1 &13.8  \\
  \textbf{T2T-ViT$_t$-14} &Transformer&2&64&64&14 &384 &1152  & 21.5 & 6.1\\
 \midrule
  T2T-ViT-7 &Performer&2&64 &64 &8 &256 &512 &4.2 &1.1\\
  T2T-ViT-12 &Performer&2&64 &64 &12 &256 &512 &6.8 &1.8\\
 \bottomrule
\end{tabular}
\label{tab: T2T_vit_structure}
\end{center}
\vspace{-22pt}
\end{table*}

As many channels in the backbone of vanilla ViT are invalid (Fig.~\ref{fig:feature_visualization}), we plan to find an efficient backbone for our T2T-ViT to reduce the redundancy and improve the feature richness.
Thus we explore different architecture designs for ViT and borrow some designs from CNNs to improve the backbone efficiency and enhance the richness of the learned features. 
As each transformer layer has skip connection as ResNets, a straightforward idea is to apply dense connection as DenseNet~\cite{huang2017densely} to increase the connectivity and feature richness, or apply Wide-ResNets or ResNeXt structure to change the channel dimension and head number in the backbone of ViT.
We explore five architecture designs from CNNs to ViT: 
\begin{enumerate} 
    \item Dense connection as DenseNet~\cite{huang2017densely};
    \item Deep-narrow vs. shallow-wide structure as in Wide-ResNets~\cite{zagoruyko2016wide};
    \item Channel attention as Squeeze-an-Excitation (SE) Networks~\cite{hu2018squeeze};
    \item More split heads in multi-head attention layer as ResNeXt~\cite{xie2017aggregated};
    \item Ghost operations as GhostNet~\cite{han2020ghostnet}.
\end{enumerate}
The details of these structure designs in ViT are given in the appendix.
We conduct extensive experiments on the structures transferring in Sec.~\ref{exp:from_cnn_to_vit}. 
We empirically find that 1) by adopting a deep-narrow structure that simply decreases channel dimensions to reduce the redundancy in channels and increase layer depth to improve feature richness in ViT, both the model size and MACs are decreased but performance is improved; 2) the channel attention as SE block also improves ViT but is less effective than using the deep-narrow structure.

Based on these findings, we design a deep-narrow architecture for our T2T-ViT backbone.
Specifically, it has a small channel number and a hidden dimension $d$ but more layers $b$. 
For tokens with fixed length $T_{f}$ from the last layer of T2T module, we concatenate a class token to it and then add Sinusoidal Position Embedding (PE) to it, the same as ViT to do classification:
\begin{equation}
\begin{aligned}
 &T_{f_0} = [t_{cls}; T_{f}] + E, &  E\in \mathbb{R}^{(l+1)\times d}\\
 &T_{f_i} = \text{MLP}(\text{MSA}(T_{f_i-1})), & i=1...b\\
 &y = \text{fc}(\text{LN}(T_{f_b}))
\end{aligned}
\end{equation}
where $E$ is Sinusoidal Position Embedding, LN is layer normalization, \text{fc} is one fully-connected layer for classification and $y$ is the output prediction.

\subsection{T2T-ViT Architecture}
\label{sec:network_structure}

The T2T-ViT has two parts: the Tokens-to-Token (T2T) module and the T2T-ViT backbone (Fig.~\ref{fig:network_structure}). 
There are various possible design choices for the T2T module. Here, we set $n=2$ as shown in Fig.~\ref{fig:network_structure}, which means there is $n+1=3$ soft split and $n=2$ re-structurization in T2T module. The patch size for the three soft splits is $P=[7,3,3]$, and the overlapping is $S=[3,1,1]$, 
which reduces size of the input image from $224\times224$ to $14\times14$ according to Eqn.~(\ref{eqn:solf_split}).

The T2T-ViT backbone takes tokens with fixed length from the T2T module as input, the same as ViT; but has a deep-narrow architecture design with smaller hidden dimensions (256-512) and MLP size (512-1536) than ViT.
For example, T2T-ViT-14 has 14 transformer layers in T2T-ViT backbone with 384 hidden dimensions, while ViT-B/16 has 12 transformer layers and 768 hidden dimensions, which is 3x larger than T2T-ViT-14 in parameters and MACs.

To fairly compare with common hand-designed CNNs, we make T2T-ViT models have comparable size with ResNets and MobileNets. Specifically, we design three models:  T2T-ViT-14,  T2T-ViT-19 and T2T-ViT-24 of comparable parameters with ResNet50, ResNet101 and ResNet152 respectively.
To compare with small models like MobileNets, we design two lite models: T2T-ViT-7, T2T-ViT-12 with comparable model size with MibileNetV1 and MibileNetV2. 
The two lite TiT-ViT have no special designs or tricks like efficient convolution~\cite{mikolov2013efficient} and simply reduce the layer depth, hidden dimension, and MLP ratio. 
The network details are summarized in Tab.~\ref{tab: T2T_vit_structure}. 
\vspace{-3pt}

\section{Experiments}
We conduct the following experiments with T2T-ViT for image classification on ImageNet. 
a) We validate the T2T-ViT by training from scratch on ImageNet and compare it with some common convolutional neural networks such as ResNets and MobileNets of comparable size; we also transfer the pretrained T2T-ViT to downstream datasets such as CIFAR10 and CIFAR100 (Sec.~\ref{exp:classification}).
(b) We compare five T2T-ViT backbone architecture designs inspired from CNNs (Sec.~\ref{exp:from_cnn_to_vit}). (c) We conduct ablation study to demonstrate effects of the T2T module and the deep-narrow architecture design of T2T-ViT (Sec.~\ref{exp:ablation}). 
\vspace{-6pt}

\subsection{T2T-ViT on ImageNet}
\label{exp:classification}
All experiments are conducted on ImageNet dataset~\cite{deng2009imagenet}, with around 1.3 million images in training set and 50k images in validation set.
We use batch size 512 or 1024 with 8 NVIDIA GPUs for training.
We adopt Pytorch~\cite{paszke2019pytorch} library and Pytorch image models library (timm) \cite{rw2019timm} to implement our models and conduct all experiments.
For fair comparisons, we implement the same training scheme for the CNN models, ViT, and our T2T-ViT.
Throughout the experiments on ImageNet, we set default image size as $224\times224$ except for some specific cases on $384\times384$, and adopt some common data augmentation methods such as mixup~\cite{zhang2017mixup} and cutmix~\cite{devries2017improved,yun2019cutmix} for both CNN and ViT\&T2T-ViT model training, because ViT models need more training data to reach reasonable performance. 
We train these models for 310 epochs, using  AdamW~\cite{loshchilov2017decoupled} as the optimizer and cosine learning rate decay~\cite{loshchilov2016sgdr}. The details of experiment setting are given in appendix. 
We also use both Transformer layer and Performer layer in T2T module for our models, resulting in T2T-ViT$_t$-14/19/24 (Transformer) and T2T-ViT-14/19/24 (Performer).
\vspace{-13pt}

\paragraph{T2T-ViT vs. ViT}
We first compare performance of T2T-ViT and ViT on ImageNet. The results are given in Tab.~\ref{tab:T2T_vit_vs_ViT}. 
Our T2T-ViT is much smaller than ViT in number of parameters and MACs, yet giving higher performance.
For example, the small ViT model ViT-S/16 with 48.6M and 10.1G MACs has 78.1\% top-1 accuracy when trained from scratch on ImageNet, while our T2T-ViT$_{t}$-14 with only 44.2\% parameters and 51.5\% MACs achieves more than 3.0\% improvement (81.5\%). 
If we compare T2T-ViT$_{t}$-24 with ViT-L/16, the former reduces parameters and MACs around 500\% but achieves more than 1.0\% improvement on ImageNet.
Comparing T2T-ViT-14 with DeiT-small and DeiT-small-Distilled, our T2T-ViT can achieve higher accuracy without large CNN models as teacher to enhance ViT. We also adopt higher image resolution as 384$\times$384 and get 83.3\% accuracy by our T2T-ViT-14$\uparrow$384.
\vspace{-13pt}

\begin{table}[]
\begin{center}
\fontsize{8pt}{12pt}\selectfont
\caption{Comparison between T2T-ViT and ViT by training from scratch on ImageNet. }
\begin{tabular}{l|ccc}
 \toprule
 Models & Top1-Acc (\%) & \tabincell{c}{Params\\(M)} &\tabincell{c}{MACs\\(G)} \\
 \midrule
 ViT-S/16~\cite{dosovitskiy2020image} & 78.1  & 48.6 & 10.1\\
 DeiT-small~\cite{touvron2020training} & 79.9 & 22.1 & 4.6\\
 DeiT-small-Distilled~\cite{touvron2020training} & 81.2 & 22.1 & 4.7\\
 \textbf{T2T-ViT-14} &\textbf{81.5}  & 21.5 & 4.8\\
 \textbf{T2T-ViT-14$\uparrow$384} &\textbf{83.3}  & 21.5 & 17.1\\
 \midrule
 ViT-B/16~\cite{dosovitskiy2020image} & 79.8 & 86.4 & 17.6\\
 ViT-L/16~\cite{dosovitskiy2020image} & 81.1 & 304.3 & 63.6\\
 \textbf{T2T-ViT-24} &\textbf{82.3} & \textbf{64.1} & \textbf{13.8}\\
 \bottomrule
\end{tabular}
\label{tab:T2T_vit_vs_ViT}
\end{center}
\vspace{-15pt}
\end{table}

\paragraph{T2T-ViT vs. ResNet} 
For fair comparisons, we set up three T2T-ViT models that have similar model size and MACs with ResNet50, ResNet101 and ResNet152.
The experimental results are given in Tab.~\ref{tab:t2t_vit_vs_resnet}. 
The proposed T2T-ViT achieves 1.4\%-2.7\% performance gain over ResNets with similar model size and MACs. 
For example, compared with ResNet50 of 25.5M parameters and 4.3G MACs, our T2T-ViT-14 have 21.5M parameters and 4.8G MACs obtain 81.5\% accuracy on ImageNet.
\vspace{-13pt}

\begin{table}[]
\begin{center}
\fontsize{8pt}{12pt}\selectfont
\caption{Comparison between our T2T-ViT with ResNets on ImageNet. T2T-ViT$_{t}$-14:  using Transformer in T2T module. T2T-ViT-14: using Performer in T2T module. * means we train the model with our training scheme for fair comparisons.}
\begin{tabular}{l|ccc}
 \toprule
 Models & Top1-Acc (\%) & \tabincell{c}{Params\\(M)} & \tabincell{c}{MACs\\(G)} \\
 \midrule
 ResNet50~\cite{he2016deep} & 76.2 & 25.5 & 4.3  \\
 ResNet50* &79.1 & 25.5 & 4.3  \\
 \textbf{T2T-ViT-14} &\textbf{81.5} & 21.5 &4.8  \\
  \textbf{T2T-ViT$_{t}$-14} &\textbf{81.7}  & 21.5 & 6.1\\
 \midrule
 ResNet101~\cite{he2016deep} & 77.4 & 44.6 & 7.9  \\
 ResNet101* & 79.9& 44.6 & 7.9  \\
 \textbf{T2T-ViT-19} & \textbf{81.9} & 39.2 &8.5 \\
 \textbf{T2T-ViT$_t$-19} & \textbf{82.2} & 39.2 &9.8 \\
 \bottomrule
 ResNet152~\cite{he2016deep} &78.3  & 60.2 & 11.6 \\
 ResNet152* &80.8 & 60.2 & 11.6 \\
 \textbf{T2T-ViT-24} &\textbf{82.3} & 64.1 &13.8 \\
 \textbf{T2T-ViT$_{t}$-24} &\textbf{82.6}  & 64.1 & 15.0\\
 \bottomrule
\end{tabular}
\label{tab:t2t_vit_vs_resnet}
\end{center}
\vspace{-20pt}
\end{table}

\paragraph{T2T-ViT vs. MobileNets} The T2T-ViT-7 and T2T-ViT-12 have similar model size with MobileNetV1~\cite{howard2017mobilenets} and MobileNetV2~\cite{sandler2018mobilenetv2}, but achieve comparable or higher performance than MobileNets (Tab.~\ref{tab:t2t_vs_mobile}). For example, Our T2T-ViT-12 with 6.9M parameters achieves 76.5\% top1 accuracy, which is higher than MobileNetsV2$_{1.4x}$ by 0.9\%.
But we also note the MACs of our T2T-ViT are still larger than MobileNets because of the dense operations in Transformers. 
However, there are no special operations or tricks like efficient convolution~\cite{mikolov2013efficient, sandler2018mobilenetv2} in current T2T-ViT-7 and T2T-ViT-12, and we only reduce model size by reducing the hidden dimension, MLP ratio and depth of layers, indicating T2T-ViT is also very promising as a lite model. We also apply knowledge distillation on our T2T-ViT as the concurrent work DeiT~\cite{touvron2020training}
and find that our T2T-ViT-7 and T2T-ViT-12 can be further improved by distillation. 
Overall, the experimental results show, our T2T-ViT can achieve superior performance when it has mid-size as ResNets and reasonable results when it has a small model size as MobileNets. 
\vspace{-13pt}

\begin{table}[]
\begin{center}
\fontsize{8pt}{12pt}\selectfont
\caption{Comparison between our lite T2T-ViT with MobileNets. Models with '-Distilled' are taught by teacher model with the method as DeiT~\cite{touvron2020training}.}
\begin{tabular}{l|ccc}
 \toprule
 Models & Top1-Acc (\%) & \tabincell{c}{Params\\(M)} & \tabincell{c}{MACs\\(G)} \\
 \midrule
 MobileNetV1 1.0x* &70.8 & 4.2 &  0.6 \\
 \textbf{T2T-ViT-7}  &\textbf{71.7} & 4.3 &1.1 \\
 \textbf{T2T-ViT-7-Distilled}  &\textbf{73.1} & 4.3 &1.1 \\
 \midrule
 MobileNetV2 1.0x* &72.8 & 3.5 & 0.3 \\
 MobileNetV2 1.4x* &75.6 & 6.9 &  0.6\\
 MobileNetV3 (Searched)  &75.2 & 5.4 &  0.2\\
 \textbf{T2T-ViT-12} &\textbf{76.5} & 6.9 &1.8 \\
 \textbf{T2T-ViT-12-Distilled} &\textbf{77.4} & 6.9 &1.9 \\
 \bottomrule
\end{tabular}
\label{tab:t2t_vs_mobile}
\end{center}
\vspace{-15pt}
\end{table}

\paragraph{Transfer learning} We transfer our pretrained T2T-ViT to downstream datasets such as CIFAR10 and CIFAR100. We finetune the pretrained T2T-ViT-14/19 with 60 epochs by using SGD optimizer and cosine learning rate decay.
The results are given in Tab.~\ref{tab:cifar}. We find that our T2T-ViT can achieve higher performance than the original ViT with smaller model sizes on the downstream datasets. 
\vspace{-5pt}

\begin{table}[]
\begin{center}
\fontsize{8pt}{12pt}\selectfont
\caption{The results of fine-tuning the pretrained T2T-ViT to downstream datasets: CIFAR10 and CIFAR100.}
\begin{tabular}{l|cccc}
 \toprule
 Models & Params (M) & ImageNet & CIFAR10 & CIFAR100  \\
 \midrule
 ViT/S-16 &48.6 &78.1 & 97.1 &87.1  \\
 \textbf{T2T-ViT-14}  &21.5 &81.5 & 97.5 &88.4 \\
 \textbf{T2T-ViT-19}  &39.1 &81.9 & 98.3 &89.0 \\
 \bottomrule
\end{tabular}
\label{tab:cifar}
\end{center}
\vspace{-31pt}
\end{table}

\begin{table*}
\small
\begin{center}
\fontsize{7pt}{12pt}\selectfont
\caption{Transfer of some common designs in CNN to ViT\&T2T-ViT, including DenseNet, Wide-ResNet, SE module, ResNeXt, Ghost operation. The same color means the correspond transfer. All models are trained from scratch on ImageNet. * means we reproduce the model with our training scheme for fair comparisons.}
\begin{tabular}{c|l|c|c|c|c|c}
 \toprule
 Model Type& Models & Top1-Acc (\%) & Params (M) & MACs (G)  &Depth & Hidden\_dim \\
 \midrule
 \multirow{3}{*}{Traditional CNN} 
 &AlexNet~\cite{krizhevsky2017imagenet} & 56.6 & 61.1 & 0.77 &- &-\\
 &VGG11~\cite{simonyan2014very}   & 69.1  & 132.8 & 7.7 &11 &-\\
 &Inception v3~\cite{szegedy2016rethinking} &77.4 &27.2 & 5.7 &- &-\\
 \midrule
\multirow{7}{*}{Skip-connection CNN} 
&ResNet50~\cite{he2016deep} & 76.2 & 25.6 & 4.3  &50 &-\\
&{ResNet50}* (Baseline) & 79.1 & 25.6 & 4.3 &50 &-\\
&\textcolor[RGB]{100, 10, 10}{Wide-ResNet18x1.5}* & 78.0 \color{blue}(-1.1) & 26.0 & 4.1 &18 &-\\
&\textcolor[RGB]{0, 0, 100}{DenseNet201}* &77.5 \color{blue}(-1.6) &20.1 & 4.4 &201 &-\\
&{\color{red}SENet50}* &80.3 \color{blue}(+1.2) & 28.1 & 4.9 &50 &-\\
&{\color{cyan}ResNeXt50}* &79.9 \color{blue}(+0.8)  & 25.0& 4.3  &50 &-\\
&{\color{magenta}ResNet50-Ghost}* & 76.2 \color{blue}(-2.9) & 19.9 & 3.2 &50 &-\\
 \midrule
\multirow{7}{*}{CNN to ViT} 
&{ViT-S/16} (Baseline) &78.1  & 48.6 &10.1  &8 &768\\
&\textbf{\textcolor[RGB]{100, 10, 10}{ViT-DN}}   &79.0 \color{blue}(+0.9) &24.5 &5.5 &16 &384 \\
&\textbf{\textcolor[RGB]{100, 10, 10}{ViT-SW}}  &69.9 \color{blue}(-8.2) & 47.9 & 9.9 &4 &1024 \\
&\textcolor[RGB]{0, 0, 100}{ViT-Dense}   &76.8 \color{blue}(-1.3) &46.7 &9.7 &19 &128-736 \\
&{\textbf{\color{red}ViT-SE}}    &\textbf{78.4} \color{blue}(+0.3) &49.2 &10.2 &8 &768 \\
&{\color{cyan}ViT-ResNeXt} &78.0 \color{blue}(-0.1) &48.6 &10.1 &8 &768   \\
&{\color{magenta}ViT-Ghost} &73.7 \color{blue}(-4.4) &32.1 &6.9 &8 &768 \\
 \midrule
\multirow{6}{*}{CNN to T2T-ViT} 
&{T2T-ViT-14} (Baseline)         &81.5 & 21.5 & 4.8  &14 &384 \\
&\textbf{\textcolor[RGB]{100, 10, 10}{T2T-ViT-Wide}}  & 77.9 \color{blue}(-3.4) & 25.1 & 5.0 &14 &768 \\
&\textcolor[RGB]{0, 0, 100}{T2T-ViT-Dense}   & 80.6 \color{blue}(-1.1) & 23.7 & 5.5  &19 &128-584 \\
&\textbf{{\color{red} T2T-ViT-SE}}     & \textbf{81.6} \color{blue}(+0.1)& 21.9 & 4.9  &14 &384 \\
&{\color{cyan} T2T-ViT-ResNeXt} & 81.5 \color{blue}(+0.0) & 21.5 & 4.8  &14 &384 \\
&{\color{magenta} T2T-ViT-Ghost}  & 79.5 \color{blue}(-2.0)& 16.3 & 3.7   &14 &384 \\
\bottomrule
\end{tabular}
\label{tab:cnn2vit}
\end{center}
\vspace{-22pt}
\end{table*}

\subsection{From CNN to ViT}
\label{exp:from_cnn_to_vit}

To find an efficient backbone for vision transformers, we experimentally apply DenseNet structure, Wide-ResNet structure (wide or narrow channel dimensions), SE block (channel attention), ResNeXt structure (more heads in multihead attention), and Ghost operation from CNN to ViT.
The details of these architecture designs are given in the appendix.
From experimental results on ``CNN to ViT'' in Tab.~\ref{tab:cnn2vit}, we can find both SE (ViT-SE) and Deep-Narrow structure (ViT-DN) benefit the ViT but the most effective structure is deep-narrow structure, which decreases model size and MACs nearly 2x and brings 0.9\% improvement on the baseline model ViT-S/16. 

We further apply these structures from CNN to our T2T-ViT, and conduct experiments on ImageNet under the same training scheme.
We take ResNet50 as the baseline for CNN, ViT-S/16 for ViT, and T2T-ViT-14 for T2T-ViT. 
All experimental results are given in Tab.~\ref{tab:cnn2vit}, and those on CNN and ViT\&T2T-ViT are marked with the same colors. 
We summarize the effects of each CNN-based structure below.
\vspace{-14pt}
\paragraph{Deep-narrow structure benefits ViT:} 
The models ViT-DN (Deep-Narrow) and ViT-SW (Shallow-Wide) in Tab.~\ref{tab:cnn2vit} are two opposite designs in channel dimension and layer depth, where ViT-DN has 384 hidden dimensions and 16 layers and ViT-SW has 1,024 hidden dimensions and 4 layers.
Compared with the baseline model ViT-S/16 with 768 hidden dimensions and 8 layers, shallow-wide model ViT-SW has 8.2\% decrease in performance while ViT-DN with only half of model size and MACs achieve 0.9\% increase. 
These results validate our hypothesis that vanilla ViT with shallow-wide structure is redundant in channel dimensions and limited feature richness with shallow layers. 

\vspace{-14pt}
\paragraph{Dense connection hurts performance of both ViT and T2T-ViT:} 
Compared with the ResNet50, DenseNet201 has smaller parameters and comparable MACs, while it has higher performance.
However, the dense connection can hurt performance of ViT-Dense and T2T-ViT-Dense (dark blue rows in Tab.~\ref{tab:cnn2vit}).

\vspace{-14pt}
\paragraph{SE block improves both ViT and T2T-ViT:} From red rows in Tab.~\ref{tab:cnn2vit}, we can find SENets, ViT-SE and T2T-ViT-SE are higher than the corresponding baseline. The SE module can improve performance on both CNN and ViT, which means applying attention to channels benefits both CNN and ViT models.

\vspace{-14pt}
\paragraph{ResNeXt structure has few effects on ViT and T2T-ViT:} ResNeXts adopt multi-head on ResNets, while Transformers are also multi-head attention structure. 
When we adopt more heads like 32, we can find it has few effects on performance (red rows in Tab~\ref{tab:cnn2vit}). 
However, adopting a large number of heads makes the GPU memory large, which is thus unnecessary in ViT and T2T-ViT.

\vspace{-14pt}
\paragraph{Ghost can further compress model and reduce MACs of T2T-ViT:} Comparing experimental results of Ghost operation (magenta row in Tab.~\ref{tab:cnn2vit}), the accuracy decreases 2.9\% on ResNet50, 2.0\% on T2T-ViT, and 4.4\% on ViT.
So the Ghost operation can further reduce the parameters and MACs of T2T-ViT with smaller performance degradation than ResNet. But for the original ViT, it would cause more decrease than ResNet. 

Besides, for all five structures, the T2T-ViT performs better than ViT, which further validates the superiority of our proposed T2T-ViT. 
And we also wish this study of transferring CNN structure to ViT can motivate the network design of Transformers in vision tasks.

\subsection{Ablation study}
\vspace{-5pt}

\label{exp:ablation}
\begin{table}[]
\begin{center}
\fontsize{7pt}{12pt}\selectfont
\caption{Ablation study results on T2T module, Deep-Narrow(DN) structure.}
\begin{tabular}{l|lccc}
 \toprule
 Ablation type& Models & \tabincell{c}{Top1-Acc\\(\%)} & \tabincell{c}{Params\\(M)} & \tabincell{c}{MACs\\(G)} \\
 \midrule
 \multirow{4}{*}{T2T module} 
 & T2T-ViT-14$_{wo\ T2T}$ &79.5 &21.1 &4.2 \\
 & T2T-ViT-14 &81.5 \color{blue}(+2.0) &21.5 & 4.8\\
 & T2T-ViT$_{t}$-14 &81.7 \color{blue}(+2.2) &21.5 & 6.1\\
 & T2T-ViT$_{c}$-14 &80.8 \color{blue}(+1.3) &21.3 & 4.6\\
  \midrule
 \multirow{2}{*}{\tabincell{c}{DN Structure}}
 & T2T-ViT-14      &81.5 &21.5 & 4.8\\
 &T2T-ViT-d768-4   &78.8 \color{blue}(-2.7)  &25.0  &5.4  \\
 \bottomrule
\end{tabular}
\label{tab:ablation_stu}
\end{center}
\vspace{-27pt}
\end{table}

To further identify effects of T2T module and deep-narrow structure, we do ablation study on our T2T-ViT.
\vspace{-12pt}
\paragraph{T2T module} To verify the effects of the proposed T2T module, we experimentally compare three different models:  T2T-ViT-14, T2T-ViT-14$_{wo\ T2T}$, and T2T-ViT$_t$-14, where T2T-ViT-14$_{wo\ T2T}$ has the same T2T-ViT backbone but without T2T module. 
We can find with similar model size and MACs, the T2T module can improve model performance by 2.0\%-2.2\% on ImageNet. 

As the soft split in T2T module is similar to convolution operation without convolution filters, we also replace the T2T module by 3 convolution layers with kernel size (7,3,3), stride size (4,2,2) respectively.
Such a model with convolution layers to build T2T module is denoted as T2T-ViT$_{c}$-14.
From Tab.~\ref{tab:ablation_stu}, we can find the T2T-ViT$_{c}$-14 is worse than T2T-ViT-14 and T2T-ViT$_t$-14 by 0.5\%-1.0\% on ImageNet. 
We also note that the T2T-ViT$_{c}$-14 is still higher than T2T-ViT-14$_{wo\ T2T}$, as the convolution layers in the early stage can also model the structure information.
But our designed T2T module is better than the convolution layers as it can model both the global relation and the structure information of the images.

\vspace{-15pt}
\paragraph{Deep-narrow structure} We use the deep-narrow structure with fewer hidden dimensions but more layers, rather than the shallow-wide one in the original ViT. 
We compare the T2T-ViT-14 and T2T-ViT-d768-4 to verify its effects.
T2T-ViT-d768-4 is a shallow-wide structure with hidden dimension of 768 and 4 layers, with similar model size and MACs as T2T-ViT-14.
From Tab.~\ref{tab:ablation_stu}, we can find after changing our deep-narrow to shallow-wide structure, the T2T-ViT-d768-4 has 2.7\% decrease in top-1 accuracy,  validating deep-narrow structure is crucial for T2T-ViT.

\vspace{-5pt}
\section{Conclusion}
\vspace{-5pt}
In this work, we propose a new T2T-ViT model that can be trained from scratch on ImageNet and achieve comparable or even better performance than CNNs.
T2T-ViT effectively models the structure information of images and enhances feature richness,  overcoming limitations of ViT.
It introduces the novel tokens-to-token (T2T) process to progressively tokenize images to tokens and structurally aggregate tokens.
We also explore various architecture design choices from CNNs for improving T2T-ViT performance, and empirically find the deep-narrow architecture performs better than the shallow-wide structure. 
Our T2T-ViT achieves superior performance to ResNets and comparable performance to MobileNets with similar model size when trained from scratch on ImageNet.
It paves the way for further developing transformer-based models for vision tasks.

{\small
\bibliographystyle{ieee}
\bibliography{egbib}
}

\end{document}